\documentclass[letterpaper, 10 pt, conference]{ieeeconf}  

\IEEEoverridecommandlockouts                              

\usepackage{graphics} 
\usepackage{epsfig} 
\usepackage{mathptmx} 
\usepackage{times} 
\usepackage{amsmath} 
\usepackage{amssymb}  
\usepackage{multirow}
\usepackage{colortbl}
\usepackage{array}
\usepackage{booktabs}
\usepackage{tabularx}

\title{\LARGE \bf
Service Robots in a Bakery Shop: A Field Study
}

\author{Sichao Song$^{1}$ and Baba Jun$^{2}$ and Junya Nakanishi$^{3}$ and Yuichiro Yoshikawa$^{4}$ and Hiroshi Ishiguro$^{5}$ 
\thanks{*This work was supported by JST Moonshot R\&D Grant Number JP- MJMS2011.}
\thanks{$^{1}$Sichao Song is with CyberAgent Inc., Tokyo, Japan
        {\tt\small song\_sichao@cyberagent.co.jp}}%
\thanks{$^{2}$Baba Jun is with CyberAgent Inc., Tokyo, Japan
        {\tt\small baba\_jun@cyberagent.co.jp}}%
\thanks{$^{3}$Junya Nakanishi is with Osaka University, Osaka, Japan
        {\tt\small nakanishi@irl.sys.es.osaka-u.ac.jp}}%
\thanks{$^{4}$Yuichiro Yoshikawa is with Osaka University, Osaka, Japan
        {\tt\small yoshikawa@irl.sys.es.osaka-u.ac.jp}}%
\thanks{$^{5}$Hiroshi Ishiguro is with Osaka University, Osaka, Japan
        {\tt\small ishiguro@irl.sys.es.osaka-u.ac.jp}}%
}

\begin{document}

\maketitle
\thispagestyle{empty}
\pagestyle{empty}

\begin{abstract}

In this paper, we report on a field study in which we employed two service robots in a bakery store as a sales promotion. Previous studies have explored public applications of service robots public such as shopping malls. However, more evidence is needed that service robots can contribute to sales in real stores. Moreover, the behaviors of customers and service robots in the context of sales promotions have not been examined well. Hence, the types of robot behavior that can be considered effective and the customers’ responses to these robots remain unclear. To address these issues, we installed two tele-operated service robots in a bakery store for nearly 2 weeks, one at the entrance as a greeter and the other one inside the store to recommend products. The results show a dramatic increase in sales during the days when the robots were applied. Furthermore, we annotated the video recordings of both the robots’ and customers' behavior. We found that although the robot placed at the entrance successfully attracted the interest of the passersby, no apparent increase in the number of customers visiting the store was observed. However, we confirmed that the recommendations of the robot operating inside the store did have a positive impact. We discuss our findings in detail and provide both theoretical and practical recommendations for future research and applications.


\end{abstract}

\section{INTRODUCTION}

Recently, social robots have been increasingly discussed. Because they are capable of communicating and interacting with people, they are expected to be useful in application scenarios such as information and guidance \cite{c1}, healthcare \cite{c2}, education \cite{c3}, and hotel service \cite{c4}. Such robots can be adopted  to maintain a certain high level of performance, generate interest, and drive engagement. Therefore, social robots can be particularly beneficial in the service industry, and such robots are referred to as service robots. According to \cite{c5,c6}, service robots can be defined as those that ``operate semi- or fully autonomously to perform services useful of the well-being of humans and equipment, excluding manufacturing operations." Often, this type of robot may possess a humanlike appearance and communicate with people in natural language \cite{c7,c8,c9,c10}.


Among the many applications in the service industry, sales promotion is an important objective for service robots \cite{c6,c12,c16,c17}. Retailers continuously search for innovative ways to engage with customers. As a novel form of interaction media, robots can be particularly effective in attracting interest and attention from potential customers, engaging with them, and facilitating their decisions to buy \cite{c6}. Previous research has explored the usage of service robots for many purposes, such as flyer distribution \cite{c17} and product recommendation \cite{c12,c16}. Regarding sales promotion and product recommendation, researchers have explored the effectiveness of service robots through field studies in robots were installed in shopping malls and stores. For example, \cite{c12} used an android salesperson to sell goods in a department store. Their robot managed to serve nearly twice as many customers as the human clerk. In a more recently study \cite{c6}, the authors placed a robot named Pepper in front of a chocolate store located at an international airport. They examined the sequential effects in a structured way, considering an entire process from passersby stopping to the final completion of purchases. In addition, they compared the robot with a tablet service kiosk and found that it was more effective.


\begin{figure*}[t]
\centering
\includegraphics[width=0.9\linewidth]{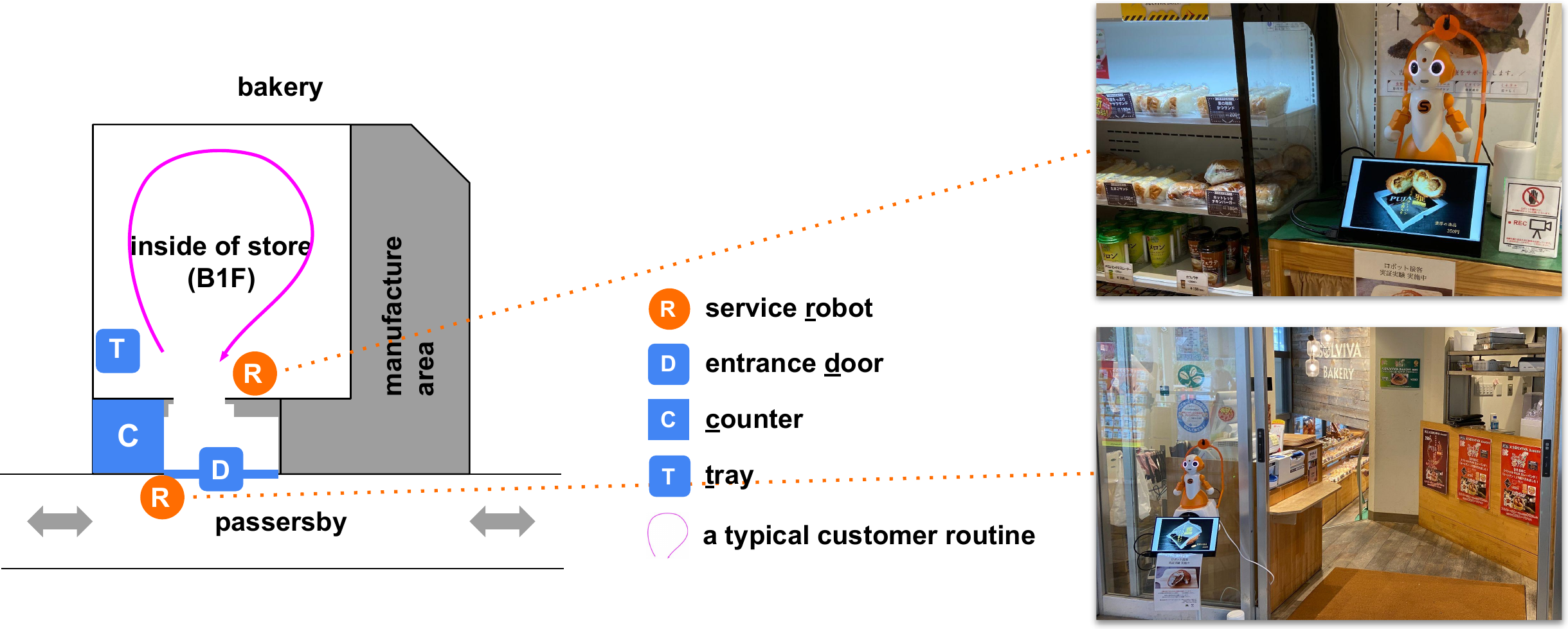}
\caption{Bakery store layout and the settings of the experiment.}\label{fig:fieldstudysettings}
\end{figure*}

Unfortunately, despite the growing interest in sales promotion, methods of application and their potential benefits in the retail industry remain unclear. Further empirical evidence remains necessary to show the effectiveness of service robots on promotions and product recommendation. Moreover, as \cite{c6} noted, existing studies are primarily more exploratory in nature. Hence, relatively little information is available on which types of robot behavior are effective and how customers respond accordingly. To address these, we performed a field study in a bakery store for almost two weeks. We aimed to (1) confirm whether the adoption of service robots could increase sales, and (2) observing the behaviors of both the robots and customers in detail to examine the effectiveness of the behaviors of the robots and frame applicable knowledge for future research and applications.


In our study, we placed two tele-operated service robots in the bakery. One was located outside at the entrance to welcome passersby and announce product information to them, and the other was placed inside the store to recommend specific products (breads) to customers (Figure~\ref{fig:fieldstudysettings}). The robots were tele-operated and thus were capable of natural conversation in the relatively noisy environment. The field experiment lasted for 11 days in total, of which the robots were in service for 8 days (Thursday $\times$ 2, Friday $\times$ 2, Saturday $\times$ 2 and Sunday $\times$ 2), whereas they were concealed and deactivated during the remaining 3 days (Monday, Tuesday, and Wednesday). We recorded a dramatic increase in sales during the period when the robots were adopted, whereas no change in sales was apparent during the period when the robots were concealed. In addition, we annotated the video recordings of the two robots. We found that the robot placed at the entrance of the store succeed in attracting the interest and attention of passersby but was not able to persuade many of them go into the store. In contrast the robot placed inside the store was found to be highly effective in recommending products and persuading the customers to purchase them. Interestingly, we discovered that a direct approach in which the robot asked customers to consider a specific bread and placed it onto a tray for them was particularly effective. The robot' service also seemed to influence the customers who visited the store with companions rather more than those who were alone.

The contributions of this work are summarized as follows. (I) We present empirical evidence that service robots can contribute to increased sales in a retail context. (II) We analyzed the of both the robot and the customers to identify those that were particularly effective. Finally (III) we discuss the roles of both robots inside and outside the store, and provide recommendations on the adoption of service robots for future applications in sales promotion.

%

\section{METHOD}
\subsection{Objectives}

The objectives of this were: (1) to confirm the effectiveness of service robots in sales promotion and product recommendation, and (2) to discovering contributing factors. We therefore conducted an empirical study to observe the customers' real in-store shopping behaviors. We set the study to a period of nearly two weeks, during which we placed the robots in service for 8 days (Thursday $\times$ 2, Friday $\times$ 2, Saturday $\times$ 2 and Sunday $\times$ 2) and concealed them, deactivated, for 3 days in between (Monday, Tuesday, and Wednesday). The purpose was to observe whether the benefit of the service robots disappeared when  they were not in service. The experiment was performed in July, 2021, from the 1st to the 11th of the month.

\subsection{tele-operated Service Robot}
We have developed a tele-operated robot system which enables an operator to remotely control and talk through the robot via a web-based video calling application. The system comprised three main components, including a robot controller application, an operator interface, and a server. A small humanoid robot, named ``SOTA" (Vstone Co., Ltd) was used. SOTA is capable of rotating its body, moving its head, making hand gestures, and blinking its LED eyes. Operators can monitor the video sent from the robot and speak accordingly through the robot in real time. Additionally, the operator can control the gaze direction of the robot through an interface.

\subsection{Field: the Bakery Store}

The bakery store was located in Osaka, Japan, on a shopping street in a residential area. Despite its relatively small size of store, the shop sells over 50 types of bread. A typical routine for a normal customer might consist of entering the store, picking up a tray, looking around to peruse the bread in a clockwise direction, and then returning back to the entrance where the counter is located to complete a transaction. Because the main space of the store is below the ground level of the entrance, passersby cannot view the inside of the store from outside. Figure~\ref{fig:fieldstudysettings} demonstrates the layout of the store and the experimental setup.

\subsection{Tasks}

The overall goal was to increase sales. Two basic strategies for achieving this could include inviting more passersby to the store and persuading them to purchase more products than they otherwise might. Therefore, we utilized two service robots with different roles. Additionally, to investigate the effects of the robots on product recommendation in particular, we selected a few types of breads out of all the products of the store. Specifically, we discussed this with the owner of the store and selected 6 types of bread for which the robots would offer specific recommendations. The goal was to observe the changes in sales of these products compared with those of the others. To differentiate these products, we refer to them \emph{recommendation breads} and \emph{non-recommendation breads}. The bakery store was prepared to double the number of the recommendation breads in case they sold out earlier before close because the usual daily breads production was optimized to reduce the number of breads remaining after closed.

Specifically, the main tasks for the robot placed at the entrance of the store were to welcoming the passersby, introducing the products in general, and asking them to enter the store and shop. If a passerby stopped in front of the robot and showed interest, the robot would then offer detailed information on the recommendation breads, including, e.g., flavor and price, and invite them to enter the store and consider those and other products. In addition, the robot would also converse with passerby if they initiated an interaction. Such conversation might be either relevant or irrelevant to the bakery store and its products but was designed to be entertaining.

The main tasks for the robot placed inside the store were to recommend the selected bread products, converse with customers and answer their questions about the products, and announce particular events such as were freshly baked bread. In terms of recommendation, the robot used two strategies that presented the product and instructions to the customers. To be specific, presenting the product was a non-directed behavior in which the robot explained information about a particular product in detail, aiming to interest customers in the products and relate to them. In contrast, instructing the customers was performed as a more directed behavior, in which the robot requested specific actions of the customers, such as by requesting that they examine a particular bread and placing it into a tray in an attempt  to persuade them and shape their behavior.

We employed two robot operators for the tasks. Both of them were female, in their 20's, and had a sufficiently high level of conversation skills. One operator was a college student who participated in a drama club, while the other was a voice actor and a particularly skilled performer and communicator. We assigned the voice actor to the operate the robot inside the store because we considered that the task would require more communication, social performance, and dynamic adjustments according to the behaviors of customers, particularly for giving instructions. We assigned the college student to the robot placed at the entrance of the store. We prepared a basic set of pre-designed scripts and information on recommended products to both of the two operators for reference. In addition, for consistency with the child-like appearance of the robot, we used voice conversion and told the operators not to admit that to the customers that they were humans if asked. Such an approach has been referred to by term ``Wizard of Oz" \cite{c18,c19}.

\subsection{Evaluation}

To assess the task performance and evaluate our study objectives, we primarily relied on the following evaluations.

\begin{itemize}
\item[(1)] Sales: total sales as well as number of customers and average sale per customer.
\item[(2)] Outside-store passerby behavior: stopping rate, conversation rate,  store visit rate, etc.
\item[(3)] In-store customer behavior: behavior annotation with key actions e.g. looking at recommendation breads and putting recommendation breads in a tray, etc.
\end{itemize}

We obtained the last two years' sales data of the bakery store for analysis. Although a relatively naive method, we assessed  sales increases by comparing the average sales during the experiment with those both during the same period of days in the previous year and the period of days two weeks before the experiment started. In addition, because we set both the days in which the robot was in service (4 days $\times$ 2) and those in which it was not (3 days between the two robot-in-service days), we were able to observe how the sales varied in relation to the adoption of our service robots.  

\begin{figure*}[t]
\centering
\includegraphics[width=0.9\linewidth]{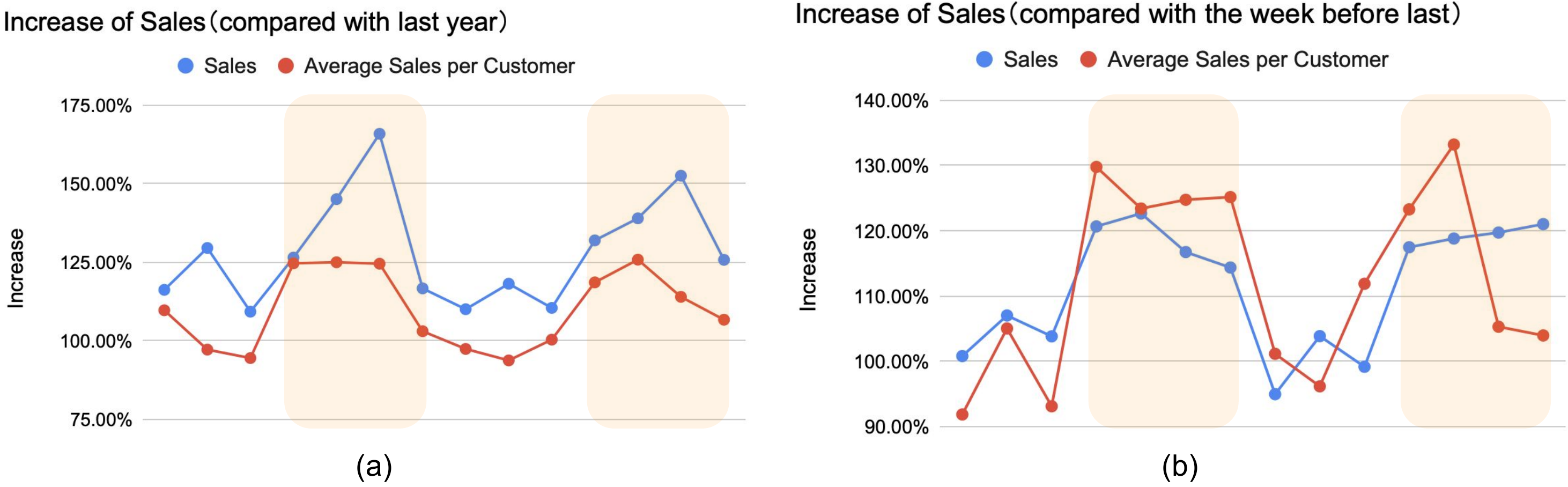}
\caption{Increase in sales and average sales per customer compared to (a) previous year and (b) the week before last. A percentage of 100.00\% represents that no change was observed with regards to the sales data. Colored areas indicate the days when the service robots were adopted.}\label{fig:increaseofsales}
\end{figure*}

We recorded a video of the experiment through a webcam on the robotic systems. With regards to the robot placed at the entrance of the store, we coded a sample of its video data (in total 4 days; Friday $\times$ 2 and Saturday $\times$ 2). With regards to the robot placed inside the store, we annotated a sample of its video data (in total 2 days; Friday $\times$ 1 and Saturday $\times$ 1) using a pre-defined codebook for both robot and the customer behaviors.


We were particularly interested in how the robot's product recommendation strategies, such as presenting products, instructing customers, , announcing freshly baked recommendation breads, or the number of different items available for sale, influenced the customers' shopping behavior. We also hoped to identify customer behaviors, e.g. looking at the robot, and attributes, e.g. visiting the store with companions, that could be positively associated with purchasing behavior towards the breads recommended by the robot, and the extent to which these applied in different situations. To annotate the videos, we followed a same procedure for each video, in which we (1) prepared a spreadsheet in which each row represented either the robot or its corresponding customers and each column represented a second in the video, and (2) labeled the robot's behaviors by referring to the codebook and put them in the same row in a time series in which the number of columns of each action label represented the length of the behavior performed, and (3) labeled each customer's behaviors by referring to the codebook, as we did for the robots, and (4) merged the action labels of the robot into each row of the customers’ in a time series in which each row represented a reference of action series that reflected the robot-customer behavior patterns. We then assembled all the action series and cleaned up the noise to obtain a large set of robot-customer behavior series from 380 customers.

We announced to all passersby and customers through a notification board that an experiment was being conducted and video was being recorded by the robot's webcam. This study was conducted on an opt-out basis for unwilling participants who chose to be removed from the video data. This experiment was approved by the facility authorities in the office building and the Research Ethics Committee of Osaka University (Reference number: R-1-5-7).

\section{RESULTS}
\subsection{Sales}

We observed an apparent increase in sales during the days when the robots were in service (Fig~\ref{fig:increaseofsales}). According to the sales data of the bakery one month before the experiment, there was an increase of the average sales per day of about 110.89\% this year (2021) compared to the previous year (2020). The overall trend of increase in sales compared to last year was confirmed by the store owner as well. However, during the experiment, we recorded a dramatic increase of the average sales per day about \textbf{138.74\%}, compared to last year, during the days the service robots were adopted, compared to only a 112.90\% increase during the days in which the robots were taken out of service. That is, although the increase in sales during the days when the robots were deactivated and concealed was in line with the overall increasing trend throughout June (one month before the experiment) and was expected, the adoption of the service robots resulted in a significant increase in sales compared to the overall trend.

In addition, we confirmed the increase of sales compared to the week preceding the previous one. The results show an increase of the average sales per day of about \textbf{119.05\%} during the days when the service robots were in service. However, we saw only a 99.33\% increase during the days when the robots were out of service.

\subsubsection{Number of Customers}

The data show an increase in the number of customers of 116.35\% this year compared to last year. During the experiment, we observed an increase of the number of customers about 116.50\% during the days the service robots were in service, and a 116.17\% increase during the days when the robots were taken out of service. In other words, both the increase in the number of customers was in line with the overall increasing trend of increase in June, regardless of whether the service robots were in use.

Additionally, we confirmed the increase in the number of customers compared to the week before last as well. There was an increase in the number of customers of about 97.99\% during the days when the service robots were adopted, and a similar 96.35\% increase during the days when the robots were removed.

\subsubsection{Average Sales per Customer}

We observed a clear increase in average sales per customer during the days when the robots were adopted (Fig~\ref{fig:increaseofsales}). There was an increase in the average sales per customer of about 95.25\% for the present year compared to the previous year. However, during the experiment, we confirmed a dramatic increase of the average sales per customer of about \textbf{117.55\%} during the days in which service robots were adopted, but only a 97.10\% increase during the days when the robots were removed from service. In other words, although the increase in the average sales per customer during the days when the robots were deactivated was in line with the overall increasing trend  throughout June and showed no difference, the adoption of the service robot led to a significant increase in the average sales per customer.

We also confirmed the increase of the average sales per customer compared to the week before last. There was a clear increase in the average sales per customer of about \textbf{120.36\%} during the days when the service robots were adopted. However, we observed only a 102.61\% increase during the days when the robots were removed.

\subsubsection{Average Number of Breads Remaining After Closed}

The average number of recommendation breads remaining after close was $3.00\pm1.41$ during the days when the service robots were adopted, significantly less than the number of $7.57\pm4.08$ during the days when the robots were removed, $t(56)=-2.70, p<.05$. In contrast, no significant difference was confirmed with regards to the average number of non-recommendation breads remaining after close, $2.32\pm1.53$, during the days when the robots were adopted compared to the number of $2.38\pm1.52$ during the days when the robots were removed, $t(56)=-0.16, p=0.87$.

\subsection{Robot at the Entrance}


We observed a total of 12461 passersby appeared in the video recordings. 

\subsubsection{Stop Rate}

The stop rate (SR) was defined as the ratio of the number of passersby who stopped at the robot and the total number of passersby. We confirmed a SR of 3.01\% for the service robot placed outside the store. 

\subsubsection{Stop Time}

The average stop time (ST) for each passerby was 47 seconds. In addition, about 23.08\% of the passersby who stopped at the robot had relatively long interaction time for over 1 minute.

\subsubsection{Conversation Rate}

The conversation rate (CR) was calculated as the number of passersby who spoke to the robot over the number of passersby who stopped at the robot. We confirmed a CR of 58.08\%. Less than half of the passersby who spoke to the robot (36.92\%) had conversations that were related to the products of the bakery store. The majority of them had conversations that were irrelevant to the breads introduced by the robot.

\begin{table}[t]
\renewcommand{\arraystretch}{1.3}
\caption{Results of logistic regression performed using the robot-customer behavior series from a number of 380 customers.}
\label{tab:logisticregression}
\centering
\begin{tabular}{>{\arraybackslash}m{4.6cm}>{\centering\arraybackslash}m{1cm}>{\centering\arraybackslash}m{1.2cm}}
\toprule
\bfseries Variables & \bfseries $P$ value & \bfseries Odds ratio (OR) \\
\hline
\bfseries Robot actions (in seconds) & & \\
\hline
Presenting recommendation breads & .66 & 1.05 \\
Instructing the customers & \bfseries .04 & 1.83 \\
Announcing freshly baked recommendation breads & .17 & 1.26 \\
Announcing remaining number of recommendation breads & .47 & 1.20 \\
\hline
\bfseries Customer actions (in seconds)  & & \\
\hline
Looking at recommendation breads & .99 & .001 \\
Looking at the robot & .42 & .97 \\
Talking to the robot & .99 & .25 \\
\hline
\bfseries Customer attributions & & \\
\hline
Sex & .92 & 1.04 \\
Child or not & .43 & 2.90 \\
With companion or not & .10 & 2.01 \\
\hline
\bfseries Interaction & & \\
\hline
Looking at recommendation breads * sex & .67 & .97 \\
Looking at the robot * sex & .41 & 1.03 \\
Talking to the robot * sex & .49 & .83 \\
Looking at recommendation breads * Child or not & .99 & 826.36 \\
Looking at the robot * Child or not & .64 & 1.02 \\
Talking to the robot * Child or not & .99 & 4.75 \\
Looking at recommendation breads * With companion or not & \bfseries .001 & 1.34 \\
Looking at the robot * With companion or not & .31 & .96 \\
Talking to the robot * With companion or not & .10 & 2.01 \\
\bottomrule
\end{tabular}
\end{table}

\subsubsection{Customer Visit Rate}

The overall customer visit rate (CVR) was defined as the ratio of the number of passersby who visited the store and the total number of passersby. We observed a CVR of 4.33\%. We also confirmed the CVR for the passersby who stopped at the robot (CVR-S). The CVR-S was calculated as the number of passersby who visited the store after stopping at the robot and the number of passersby who stopped at the robot. We confirmed a CVR-S of 16.53\%.

\subsection{Robot inside the Store}

We annotated the sample video data with regards to the service robot placed inside the bakery store and obtained a set of robot-customer behavior series from a number of 380 customers. A logistic regression was performed to confirm the effects of the robot's behaviors, the customers' behaviors, the attributions of the customers,  and the interactions between the customers' behaviors and their attributions, on the likelihood that the customers would purchase the recommendation breads or not. The logistic regression model was statistically significant, $X^2 (19, N=380)=134.64, p<.001$. The model explained 41.2\% (Nagelkerke $R^2$) of the variance in purchase behavior and correctly classified 78.7\% of cases. Table~\ref{tab:logisticregression} lists the results.

Particularly, the robot's behaviors, especially that of instructing the customers, towards the purchase of the recommendation breads was significantly influential. The customers were nearly twice as likely to purchase the recommendation breads if the robot's instructing behavior lasted 1 second longer ($p<.05, OR=1.83$). The customers that visited the store with companions also seemed to be twice as likely to purchase the recommendation breads ($p<.1, OR=2.01$) compared to those who were alone. Moreover, a significant interaction effect between the amount of time the customers looked at the recommendation breads and whether they were present with companions was discovered as well ($p<.001, OR=1.34$).

\section{DISCUSSION}
\subsection{Service Robot for Sales Promotion}

The bakery's sales data suggested a dramatic increase in sales during the days when the service robots were adopted, but no obvious increase during the days when the robots were removed from service. Therefore, our results add to the existing scarce empirical evidence of the effectiveness of service robots for sales promotion. Interestingly, we found that the number of customers per day during the field study remained mostly at the same level, regardless of presence of the service robot. The increase in average sales per customer particularly that contributed to the change in the total sales. In other words, although the same number of customers compared to both the previous year and the week before last visited the store during the period of experiment, they purchased either more products or more expensive products when the service robots were in operation.

We analyzed the video recordings of the robot placed at the entrance of the bakery store and found that it failed to convince passersby to enter the store as we expected. Indeed, we found the robot managed to attract attention from the passersby, particularly for children. However, when looked into the details of the conversations, we found  that only about a third of the passersby who talked to the robot had conversations that were relevant to the products of the bakery. The majority of them had some sort of conversations that were, for instance, about the robot itself, their own lives and hobbies, or relatively strange questions to probe the conversational ability of the robot. Therefore, the robot attracted many passersby who were particularly interested in the robot itself. They briefly interacted with the robot and then left after their curiosities were satisfied. In other words, they were not particularly concerned or interested in the robot's introduction of the store and the breads. Convincing such people to visit a store remains a difficult task.

Although the robot placed at the entrance appeared to be less effective on converting the passersby into store visitors, it helped to shape their impressions on the bakery store. We often observed that passersby stared at the robot while walking pass it. During the experiment, we recognized 9 groups of people, mostly with children, repeatedly approach the robot and had long conversation with it. We also heard that some of the customers called the bakery store ``a store with robots" instead of the store name. Such findings suggest the potential effect of utilizing service robots for branding. 


On the other hand, the analysis of the video recordings of the robot placed inside the store suggested the influence of the robot on the customers' purchase behavior. Particularly, we found that the robot's instructing behavior towards the customers was significantly associated with their purchase behavior towards the recommendation breads. If the robot's instruction towards a customer, e.g., examine the recommendation breads and place them into the tray, increased 1 second in duration, they would be nearly twice as likely to eventually purchase one or more of the breads. It should be noted that we did not confirm the immediateness of such an effect of the robot's instructing behavior. In other words, the influence of the robot seemed not be simply reflected in the customer's action that immediately followed by the start of the robot's instructing behavior. Nonetheless, the instructing behavior would help to raise a customer's attention and interest towards the recommendation breads and consequently plant a ``seed" of motivation of purchase.


We also discovered that the robot seemed to be potentially effective on the customers that visited the store with companions. Among such customers, those who were more likely to spend longer examining the recommendation breads were found to be more likely to eventually purchase them. A plausible explanation is that the presence of the service robot and the information received from it helped to form common topics among group members, which consequently promoted joint attention towards the recommendation breads and led to the purchase behavior. We examined the video recordings and observed that the customers that visited the store with companions did often have in-group conversations about the robot and the recommendation breads and seem to spend longer time in the store.

Besides, several other factors showed potential effects on the customers' purchase behavior towards the breads recommended by the robot as well. As listed in table~\ref{tab:logisticregression}, the robot seemed to be very effective on children. Indeed, children could be particularly curious and interested in novel things such as a robot. Children seemed more likely to put the breads recommended by the robot in the trays, and this was interfered from their spent time on talking to the robot and examining the recommendation breads. Although no significant probability was obtained, possibly due to the lack of data regarding children, such findings could nonetheless be valuable for sales promotion strategies for service robots.


\subsection{Implications for Future Applications}
\subsubsection{Placement of Service Robots}

In general, we suggest that it would be valuable to place service robots both outside and inside the store for the maximum effect. It is important to begin to influence the customers' impressions before they enter the store. However, each of the robot has its unique strengths. Based on our findings, we offer our suggestions on utilizing service robots for sales promotion. ($\alpha$) Consider placing a robot outside the store for branding. ($\beta$) Consider placing a robot inside the store for product recommendation, and ($\gamma$) consider placing robots both outside and inside a store to maximizing their effects on sales promotion.


\subsubsection{Recommendation Strategies}

We found that, among several different product recommendation strategies, instructing behavior of the robot was particularly effective on production recommendation. Moreover, compared to the customers that visited the store alone, those who did so with companions seemed to be affected by the robot more. Therefore, we suggest some design recommendation strategies for service robots. ($\alpha$) Make use of long (within a reasonable length of time) instructing behavior, that ``command" the customers to do something, to promote and shape the customers' shopping behaviors. ($\beta$) Target customers who visited the store with companions more. ($\gamma$) Target children as potential customers, either who visited alone or with their parents (however, ethical issues need to be concerned).

\subsection{Limitations and Future Work}

Several points could be addressed in future research. First, we adopted a child-like robot in the experiment. Because the appearance of a robot could affect its impression and persuasiveness, future studies may utilize different types of robots, even those with non-humanlike appearance. Second, although our field study lasted for over 1 week, there could still be substantial novelty effects. Therefore, long-term studies remain needed to confirm the effects of utilizing service robots for sales promotion over longer application periods. Finally, we adopted tele-operated robots in this study to achieve natural conversation with people. However, particularly for the objective of product recommendation, our findings indicate the possibility of adopting autonomous robot systems. We plan to develop and test such autonomous systems in future work.

\addtolength{\textheight}{-14cm}   

\end{document}